\documentclass[9pt, conference, final,oneside,twocolumn,letterpaper]{IEEEtran}
\usepackage{amsmath}
\usepackage{amsmath}
\usepackage{amsfonts}
\usepackage{amssymb}
\usepackage{graphicx} 
\usepackage{cite}
\usepackage{multirow}
\usepackage{array}
\usepackage{comment}
\title{Elitism Levels Traverse Mechanism For The Derivation of Upper Bounds on Unimodal Functions}
\begin{document}
\author{\IEEEauthorblockA{Aram Ter-Sarkisov}\\
\IEEEauthorblockA{Department of Computer Science\\
Massey University\\
Wellington, New Zealand\\
Email: a.ter-sarkisov@massey.ac.nz}}
\maketitle
\begin{abstract}
\noindent In this article we present an Elitism Levels Traverse Mechanism that we designed to find bounds on population-based Evolutionary Algorithms solving unimodal functions. We prove its efficiency theoretically and test it on OneMax function deriving bounds $c \mu n \log n - O(\mu n)$. This analysis can be generalized to any similar algorithm using variants of elitist selection and genetic operators that flip or swap only 1 bit in each string.   
\end{abstract}
\section{Introduction}
\noindent We analyze an elitist population-based Evolutionary Algorithm with population size $\mu$ and recombination pool size $\lambda, (\mu+\lambda)$EA using a genetic operator 1-Bit-Swap that recombines information between parents (see \cite{tersarkisov2010}).\\
\linebreak
\noindent Most research in theoretical EA community is focused on mutation-based single species algorithms such as $(1+1)_{\frac{1}{\mu}}$ EA (see e.g. \cite{droste02, doerr10, doerr113}) with some sharp bounds on runtime obtained for OneMax function such as $0.982 n \log n$ in \cite{doerr113}.\\
\linebreak 
Results on population-based algorithms are less abundant, and are restricted to mostly $(\mu + 1)_{\frac{1}{\mu}}$EA (see \cite{witt04}) with upper bound $O(\mu n +n\log n)$ and $(1 + \lambda)_{\frac{1}{\mu}}$EA (see \cite{jansen05, he2010}) with upper bound on OneMax $O(n \log n +n \lambda)$ in \cite{jansen05} and all linear functions $O(\lfloor \frac{n \log n}{\lambda} +\frac{n \log \log \lambda}{\log \lambda} \rfloor )$ in \cite{he2010}.\\
\linebreak 
Although so far $(\mu+\lambda)$ or $(N+N)$ EAs have deserved less attention, they have been the subject of analysis in \cite{heyao02, heyao04, chenhe09}. Specifically, in \cite{chenhe09} it was derived that for a $(N+N)$ EA with mutation and tournament selection solving OneMax the upper bound is $O(n N\log N + n \log n)$ if measured in the number of function evaluations.\\ 
\linebreak
Unfortunately many of these results are not directly comparable due to the difference in selection functions (fitness-proportional, truncation, elitist, tournament, etc) and elitism settings (save 1 best species or some variable proportion).\\
\linebreak
Even more significantly, it was shown already in \cite{heyao02} that population effect is generally problem-specific, so it is quite hard to generalize findings to other functions. There is ample evidence though (e.g. \cite{jansen05, witt04}) that for mutation-based algorithms (incl. Randomized Local Search, RLS) optimizing simple functions such as OneMax population is not beneficial and tends to degrade performance.\\
\section{Algorithms and Problems}
\subsection{Algorithm}
\noindent Although the mechanism described in this paper is quite universal, we test it on $(\mu+\lambda)EA_{1BS}$ solving OneMax problem. This problem is well-known in EA community, recent achievements include \cite{doerr102, doerr113} with some sharp bounds. We selected this problem due to its simplicity and the ability to compare our findings to those available already.

\begin{table}[!h]
\caption{Algorithm}
\centering
\label{tab:tab1}
\begin{tabular}{l l}
\hline
& $(\mu+\lambda)$ Evolutionary Algorithm using 1-Bit-Swap (1BS)\\
\hline
1 & create $\mu$ starting species at random\\
2 & $t=0$ \\
3 &  \textbf{loop}\\
4 &  \ \ \ select using a variant of fitness-proportional Tournament selection  \\
&    \ \ \ $\frac{\lambda}{2}$ pairs of parents into the pool\\
5 &  \ \ \ swap bits in each pair\\
6 &  \ \ \ keep the currently-elite species in the population, replace the rest \\
&    \ \ \ with the pool, first with new currently-elite, then at random\\
7 &  \ \ \ $t=t+1$\\
8 &  \textbf{end loop}\\
\hline
\end{tabular}
\end{table} 

\subsection{Selection function}
\noindent Throughout the article we analyze an elitist recombination-driven $(\mu +\lambda)$ EA using a variant of tournament selection. It is both simple to implement and analyze. But since we recombine information between parents, we are interested in forming $\textit{pairs}$ of species in the recombination pool, and on the construction of these pairs the properties of the algorithm will be derived. This formation occurs in the following way:

\begin{table}[!h]
\caption{Selection Function}
\centering
\label{tab:tab2}
\begin{tabular}{l l}
\hline
& Variant of Tournament Selection\\
\hline
1 & $k=0$\\
2 & \textbf{loop}\\ 
3 & \ \ \ select two species from the population at random\\
4 &  \ \ \ examine their fitness, the better one enters the pool\\
5 &  \ \ \ $k=k+1$ \\
6 &  \textbf{end loop}\\
\hline
\end{tabular}
\end{table} 

\noindent Thus it is obvious that better-fit species have higher chances of entering the pool, so we can expect the proportion of $\alpha$ species to be higher in the pool rather than in the population. 

\subsection{1-Bit-Swap Genetic Operator}
We apply the 1-Bit-Swap operator that was found to be useful solving a large number of test problems in \cite {tersarkisov2010} and was analyzed extensively in \cite{tersarkisov11, tersarkisov112} to have outperformed the mainstream RLS algorithm both theoretically and numerically.\\
\linebreak
Another advantage of 1BS is that we can compare it directly to RLS, since both are local search operators that cannot move too far from the current best search point. The operator works in the following way:   
\begin{table}[!h]
\caption{Genetic Operator}
\centering
\label{tab:tab3}
\begin{tabular}{l l}
\hline
& 1-Bit-Swap Operator\\
\hline
1 & $j=0$ \\
2 & \textbf{loop}\\ 
3 & \ \ \ select a bit in the first parent uniformly at random\\
4 &  \ \ \ select a bit in the second parent uniformly at random\\
5 &  \ \ \ swap values in these bits\\
5 &  \ \ \ $j=j+1$ \\
6 &  \textbf{end loop}\\
\hline
\end{tabular}
\end{table} 

\section{Definitions}
\subsection{Fitness levels partition}
\noindent Basic approach to analyzing elitist EA with a simple 1-bit mutation solving unimodal binary-encoded EAs was introduced by Wegener in \cite{wegener2003} that is based on fitness partitioning: on a set of binary strings $\{0,1 \}^n$ size $2^n$ a partition into a finite number of nonempty subsets $A_1, A_2, \cdots, A_m$ is defined with ordering $A_1 \precsim A_2 \precsim \cdots \precsim A_m$ s.t. all $a \in A_m$ are the global optimum.\\
\linebreak
This approach allows definition and derivation of the lower bound of success probability of transition between states, $s_{i}(a)=P(A_{i+1}|A_i)$ and the upper bound on expected convergence time of the algorithm, expected first hitting time of the best fitness level, $E(X_f) \leq s_{1}^{-1} + s_{2}^{-1}  \ldots + s_{m-1}^{-1}$. This idea can be extended to the situation when we apply non-negative weights $w(f)$ (see \cite{wegener2003, droste02}) and to derive lower bounds (by considering the upper bound on $s_i(a)$.\\
\linebreak
Another tool used extensively in the analysis of EAs are potential (auxiliary) functions that measure progress (see \cite{wegener2003}). This is especially useful when working on functions that have fitness plateaus (see e.g. \cite{jansen01}), in which case we make the difference:
\begin{enumerate}
\item fitness functions decide whether the new binary input (species) is better than the old one
\item potential function tracks the progress between states of the algorithm (fitness levels) 
\end{enumerate}  
OneMax (or some simple transformation of it) is used as a potential function for more complicated problems (Royal Roads, Binary Values, Short/Long Path etc). 
\subsection{Elitism Levels Partition}
\noindent In this article we extend this approach to a population-based elitist algorithm, but rather than tracking the traverse of levels of fitness, we do the same to the levels of elitism, i.e., number of elite species in the population.\\
\linebreak
We focus on species that can either evolve to the currently-best over 1 iteration or are already best. Therefore, the population is broken down into three disjoint subsets:
\begin{align*}
\alpha:& \textnormal{ currently best species } \\
\beta_1:& \textnormal{ species with next-best fitness }\\
\beta_{-1}:& \textnormal { the rest of the population that cannot evolve over 1 generation} 
\end{align*}
Since 1BS swaps exactly 1 bit between two parents, this partition in combination with the assumptions made above enables construction of a very precise model, since the value of $\alpha$ cannot 'jump` more than 1 level of fitness and only $\alpha$-species can breed better population, but only $\beta_1$-species may evolve into $\alpha$ and change the probability of evolution.
\subsection{$\alpha$-levels subpartition}
\noindent This additional partition is necessary for functions with plateaus for which we use potential functions explained above. The need for it becomes evident in the next section, when probabilities of evolving elite species on two types of functions are compared. In addition to the elitism levels partition, for functions with plateaus we need to subpartition the $\alpha$-level.\\
\linebreak
In slight abuse of notation in the rest of this article, we denote $A$ the set of chromosomes in the population with the highest fitness. Also $\gamma$ is the length of the plateau of fitness. Therefore the set A can be partitioned into
\begin{equation*}
A=A_{0} \cup A_{1} \cup \cdots \cup A_{\gamma-1}
\end{equation*}    
where each subset $A_m$ has equal fitness. In order to differentiate between $A_m$, we assign each elite species an additional auxiliary function, $V_m$ that tracks progress to the next level by counting the number of 1-bits in the fitness level:$ \ A_0 \precsim A_1 \precsim \ldots \precsim A_{\gamma-1}$ with corresponding auxiliary values $V_0 < V_1 \ldots < V_{\gamma-1}$, i.e. OneMax is used as an auxiliary function. Species with both highest fitness and auxiliary values can be viewed as super-elite or $\alpha^{\ast}$.\\
\linebreak
In the next section we use the notation $\alpha, \ \alpha^{\ast}$ to denote the set of elite or super-elite species, an element of that set and the size of it. This is done to reduce notational clutter. 

\section{Elitism Levels Traverse Mechanism for Upper Bounds}
\label{sec:main_sec}
\noindent In this section we present the main result of the article on a general function that is later confirmed by further application to OneMax Test Function. We are interested in the upper bounds on optimization time (for explanation of Landau notation see e.g. Chapter 9 in \cite{knuth95}).\\
\linebreak
\noindent The working of the Elitism Levels Traverse Mechanism can be illustrated by an example from immunology.\\
\linebreak
There exists a population of species size $N$, which is susceptible to $M$ types of infection, which are mutually exclusive, i.e. a species cannot be infected by more than one infection. The size of each set of infected species cannot be larger than  $m_j$. We denote $E^{\ast}_j$ an event that there are $ 1 \leq r \leq m_j$ infected species of type $j$, of which exactly one spawns an infected offspring that destroys a healthy member of the population. Since the sets of infected species are mutually exclusive, by additivity we obtain the probability that any of the infected species adds exactly one infected offspring:

\begin{equation*}
P \bigg( \bigcup_{j=1}^{M}E^{\ast}_{j} \bigg) = P \bigg( \bigcup_{j=1}^{M} \bigcup_{r=1}^{m_j} E^{\ast}_{jr} \bigg)  = \sum_{j=1}^{M} \sum_{r=1}^{m_j} P ( E^{\ast}_{jr} ) =  \sum_{j=1}^{M}P(E^{\ast}_j)
\end{equation*}             

\noindent This expression is quite complicated for a number of reasons, e.g. the knowledge of $m_j$. Although we can find bounds on the partial sum of rows of Pascal triangle, it is guaranteed to make the derivation quite messy. Therefore we need to lower-bound this probability. We do this by considering only one infected species of each type rather than $r$ and the event of spawning exactly one infected species by $E_j$. This gives us the lower bound on the total probability of adding exactly one infected offspring, which is proven in Appendix \ref{sec:main_eq}:

\begin{equation}
P(E^{\ast}_j) \geq P(E_j) \leftrightarrow \sum_{j=1}^{M} P(E^{\ast}_j) \geq  \sum_{j=1}^{M} P(E_j)
\label{eq:main_ineq}
\end{equation} 

\noindent In the notation of EA, $N = \frac{\lambda}{2}$, the number of pairs of parents in the recombination pool with parents that are able to produce exactly one elite offspring. $\delta \mu$  (for $0< \delta <1$) is the number of elite individuals in the population that, once it is reached, the probability to generate an offspring with higher fitness is arbitrarily close to one, i.e. $1-o(1)$. We also have $n-1$ levels of fitness. Combining this with the upper bound on the probability of adding elite offsprings to the population, we obtain the upper bound (worst-case) on the optimization time of the algorithm:

\begin{equation}     
\mathbf{E} \tau = O \bigg( \sum_{k=1}^{n} \sum_{\alpha=1}^{\delta \mu} \frac{1}{\sum_{j=1}^{M} P(E_{j} (\alpha,k))} \bigg)
\label{eq:main_eq}
\end{equation}

\noindent Derivation of the upper bound from Equation \ref{eq:main_eq} is rather versatile. We need to identify pairs of possible parents $<p_1, p_2>$ such that there exists some probability of swapping bits between parents $\varphi_{j}(k) >0$ that as a results of applying a genetic operator to this pair either a new $\alpha$ species evolves from lower-ranked ones or an existing $\alpha$  is preserved after the recombination.\\   
\linebreak
Intuitively, for the functions with plateaus both the population size and the number of elite species are more important than for those without plateaus. In the remainder of this section we show that the probability to add a super-elite offspring when solving a function with plateaus is less than the probability to add an elite offspring when solving functions without plateaus.\\   
\linebreak  
For the rest of this section we denote $f_1$ function without plateaus and $f_2$ function with plateaus. What we show is that $P(E^{f_1}_{j}(\alpha,k)) \geq P(E^{f_2}_{j}(\alpha,k))$.

\subsection{Functions without plateaus}

\noindent For this type of unimodal functions (e.g. OneMax) intuitively it is easier to add an elite offspring and thus reduce the optimization time, but we need to show it rigorously.\\  
\linebreak
The probability to select a pair with an $\alpha$-parent can be bounded by 
\begin{equation*}
P^{f_1}_{sel} \geq  \frac{\alpha}{\mu} \xi_1
\end{equation*}
where $\xi_1$ is the probability to select a non-elite species to be paired with the elite one. Also bound the probability to flip the bits $\varphi_j(k) \geq \eta \ \forall \ j,k$. So the probability of an event $E_j$ that includes pairs with elite species is 
\begin{equation*}
P(E^{f_1}_j) \geq  \binom{\frac{\lambda}{2}}{1} \frac{\alpha}{\mu} \eta \xi_1 = \frac{\lambda \alpha \eta \xi_1}{2 \mu}
\end{equation*} 
The probability to select a pair without the the currently-elite species is lower-bounded by $\frac{\rho_1}{\mu}$. By breaking down the set of parents $M_{f_1}$ in the recombination pool into those including $\alpha-$ parents,  $M^{\ast}_{f_1}$ and those that do not, $M^{\ast \ast}_{f_1}$, we can find the lower bound on the probability of adding another elite species:
\begin{equation*}
P(S^{f_1}_{\alpha,k}) \geq \frac{M^{\ast}_{f_1} \lambda \alpha \xi_1 \eta}{2 \mu} + \frac{M^{\ast \ast}_{f_1} \lambda \eta \rho_1 }{2 \mu}
\end{equation*}    
\subsection{Functions with plateaus}
\noindent As noted in \cite{chenhe09}, algorithms with well-chosen population size perform similar to, and best individuals evolve along the same path as $(1+1)$EA. The difference between ($\mu+ \lambda) \textnormal {and } (1+1)$ lies in the cost of traversing plateau. For this type of functions the length of plateau $ \gamma >1$. So we have K plateaus w.l.o.g. of the same length $\gamma \in \mathbb{Z}^{+}, \textnormal{ and } n= \gamma K$.\\
\linebreak
Also we assume that at the start of the algorithm each `bin' (plateau) starts with an equal number of 1's and 0's uniformly distributed, therefore fitness of the best species at the beginning of the run is 0. To track progress between jumps in fitness values we use OneMax as an auxiliary function (roughly along the lines of using potential or distance functions, see e.g. \cite{he2010}) that sums bits in the plateau.\\
\linebreak
The tricky part in this analysis is that the selection is based on fitness of the string rather than auxiliary function, but the progress towards the next level of fitness plateau depends on the number of parents with highest auxiliary value, $V_s \textnormal{ rather than } f(s)$. By denoting the subset of $\alpha$ with highest auxiliary function $\alpha^{\ast}$, we notice that $f(\alpha)=f(\alpha^{\ast}), \textnormal{ and } V_{\alpha} < V_{\alpha^{\ast}}$. Also trivially $\alpha^{\ast} \leq \alpha$ (for the case of functions without plateaus these functions are identical and last two expressions are equalities).\\
\linebreak
As shown before, for a unimodal function without plateaus regardless of fitness function, the probability that one of the parents is elite is $\frac{\alpha}{\mu}$, since if two elite species are selected for breeding, parent is chosen randomly. Obviously $\frac{\alpha^{\ast}}{\mu} \leq \frac{\alpha}{\mu}$. Additionally,
\begin{equation*}
\frac{\alpha^{\ast}}{\mu} \cdot \frac{(\alpha-\alpha^{\ast})}{\mu} \cdot \frac{1}{2} \leq \frac{\alpha^{\ast}}{\mu} \leq \frac{\alpha}{\mu}
\end{equation*}               
Obviously, unlike $f_1$, for the evolution process on $f_2$ only a small subset of parents are of use, these having the highest and next-highest auxiliary values. Therefore pairs that do not include at least 1 of these parents can't add an $\alpha^{\ast}$ offspring. Similar to $f_1, \ M_{f_2}=M^{\ast}_{f_2}+M^{\ast \ast}_{f_2}$ and clearly $M^{\ast}_{f_2} \leq M^{\ast}_{f_1} \textnormal{ and } M^{\ast \ast}_{f_2} \leq M^{\ast \ast}_{f_1}$.\\
\linebreak
Along the lines of arguments in the previous subsection, $\exists \xi_2$ s.t. probability to select a non super-elite parent in addition to the super-elite one is upper-bounded by it. We get: 
\begin{equation*}
P^{f_2}_{sel} \leq \frac{\alpha^{\ast}(\alpha- \alpha^{\ast}) \xi_2}{2 \mu^2}
\end{equation*}
so the probability of an event that an $\alpha^{\ast}$ parent is added to a pool and a new $\alpha^{\ast}$ offspring evolves is upper-bounded by
\begin{equation*}  
P(E^{f_2}_j) \leq \frac{M_{f_2}^{\ast} \lambda \alpha^{\ast}(\alpha -\alpha^{\ast}) \xi_2 \eta}{4 \mu^2}
\end{equation*}
where $\eta$ is the lower bound on the probability of swapping bits. Therefore the probability to add one more $\alpha^{\ast}$ species to the population is
\begin{equation*}
P(S^{f_2}_{\alpha,k}) \leq \frac{M_{f_2}^{\ast} \lambda \alpha^{\ast}(\alpha -\alpha^{\ast}) \xi_2 \eta}{4 \mu^2} + \frac{M_{f_2}^{\ast \ast} \lambda \eta \rho_2}{2 \mu} 
\end{equation*}
Combining the inequalities above, and taking $M_{f_1}^{\ast \ast} \geq M_{f_2}^{\ast \ast} + \frac{\epsilon}{\rho_1}, \ 0< \epsilon <1$, we compare the values in the first and second fractions in the expressions for $P(S^{f_1}_{\alpha,k}) \textnormal{ and } P(S^{f_2}_{\alpha,k})$. It is easy to see that $\exists \textnormal{ two constants } \psi_2 < \psi_1 \textnormal{ s.t. }$
\begin{equation}
P(S^{f_2}_{\alpha,k}) \leq \psi_2 < \psi_1 \leq P(S^{f_1}_{\alpha,k}) 
\end{equation}
\section{Upper Bound on Runtime of $(\mu + \lambda)$EA$_{1BS}$ on OneMax test function Using Elitism Levels Traverse Mechanism}
\noindent In this section we present our findings on the upper bounds on runtime of $(\mu+\lambda)$EA with 1-Bit-Swap operator optimizing OneMax function using the Elitism Levels Traverse Mechanism. We distinguish four pairs of parents that make possible evolution of currently-elite species:
\vspace{-5pt}
\begin{align*}
E_1&:<\alpha, \beta_1>\\
E_2&:<\beta_1, \beta_1>\\
E_3&:<\alpha, \beta_{-1}>\\
E_4&:<\beta_1, \beta_{-1}>\\
\end{align*}  
We do not consider the obvious pair $<\alpha, \alpha>$ as it either adds two elite offsprings, of generates an offspring with higher fitness, something we do not use in the Mechanism. \\
\linebreak
For the upper bound on optimization time we only consider increase of the number of elite species by at most one. Increase by two or more is ignored, or otherwise transformed into any of the lower-ranked species. Similar approach was used in \cite{chenhe09} in bounding the takeover time. 
\subsection{Simple upper bound}
\noindent Of these four cases we start analysis with the first two. Main reason is that the other two cases involve cubic function, which becomes quite complicated to solve (see next subsection). For the cases $E_1, E_2$ we get the following probabilities of success:
\begin{align*}
P(E_1) &= 2 \binom{\frac{\lambda}{2}}{1} \varphi_1(k) \cdot  \frac{\alpha}{\mu} \cdot\frac{\beta_1}{\mu}\Bigg(1-\frac{\alpha}{\mu}\Bigg) \\
&= \frac{\alpha \beta_1 \lambda(\mu-\alpha)\varphi_1(k)}{\mu^3}  
\end{align*}  
\begin{align*}
P(E_2)&= \binom{\frac{\lambda}{2}}{1} \varphi_2(k) \Bigg(\frac{\beta_1}{\mu}\Bigg(1-\frac{\alpha}{\mu}\Bigg)\Bigg)^2 = \frac{\lambda \varphi_2(k) \beta_{1}^2(\mu-\alpha)^2}{2 \mu^4}
\end{align*}
The probability of at least 1 of these events is 
\begin{align*}
P(S_{\alpha,k}) & \geq P(E_1)+P(E_2) = \frac{2 \varphi_1(k) \alpha \beta_1 \lambda (\mu-\alpha)}{\mu^3} \\
& + \frac{\varphi_2(k) \lambda \beta_1^2 \varphi_1(k)(\mu-\alpha)^2}{2 \mu^4}
\end{align*}
and, since $P(S)$ is minimal, the upper bound on expected time to traverse levels of elitism large enough to get a $1-o(1)$ probability of evolution is 
\begin {IEEEeqnarray}{rCl}
\mathbf{E}\tilde{T}_{\alpha,k} \leq \sum_{\alpha=1}^{\delta \mu} \frac{1}{P(S_{\alpha})}
\end{IEEEeqnarray}
The expression for the expected first hitting time we obtain as a result of this setup is 
\begin{align}
&\mathbf{E}\tilde{T}_{\alpha,k}  \leq 2 \mu^4 \sum_{\alpha=1}^{\delta \mu} \frac{1}{\beta_1 \lambda(\mu-\alpha)(2 \alpha \mu \varphi_{1}(k) - \alpha \beta_1 \varphi_{2}(k) + \beta_1 \mu \varphi_2(k))} \nonumber \\
& = \frac{2 \mu^4}{\lambda} \cdot  \nonumber \\
&\sum_{\alpha=1}^{\delta \mu}\frac{1}{(\varphi_2(k)-2 \mu \varphi_1(k))\alpha^2 +(2 \mu^2 \varphi_1(k)-2 \mu \varphi_2(k)) \alpha + \mu^2 \varphi_2(k)} \nonumber \\
&=\frac{2\mu^4}{\lambda(\varphi_2(k)-2 \mu \varphi_1(k))}\sum_{\alpha=1}^{\delta \mu}\frac{1}{\alpha^2 +b_1 \alpha +b_0}
\label{fht1}
\end{align}  
where 
\begin{align*}
b_0 &= \frac{\mu^2 \varphi_2(k)}{\varphi_2(k)-2 \mu \varphi_1(k)} \\
b_1 &= \frac{2 \mu(\mu \varphi_1(k)-\varphi_2(k))}{\varphi_2(k) -2 \mu \varphi_1(k)}
\end{align*}
At this point we set $\beta_1$ pessimistically to 1 to simplify the derivation. This a quadratic equation in $\alpha$. The full solution to Equation \ref{fht1} is in Appendix \ref{sec:eq2}.\\
\linebreak
The optimization time is 
\begin{equation}
\mathbf{E}\tau_{(\mu+\lambda)EA_{1BS}} = O(\mu^{1+\varepsilon_2} n \log n)
\end{equation}
for some constant $\varepsilon_2(\mu)$. For the second option of $\delta=\frac{c}{\mu}$ the upper bound becomes
\begin{equation}
\mathbf{E}\tau_{(\mu+\lambda)EA_{1BS}} = O(\mu n \log n)
\end{equation}
or, in the number of function evaluations
\begin{equation}
\mathbf{E}\tau_{(\mu+\lambda)EA_{1BS}} = O(\lambda \mu n \log n)
\end{equation}
\subsection{Refined upper bound}
\noindent We add the other two cases to obtain a sharper upper bound on optimization time, we set $\beta_1=1$:
\begin{align*}
P(E_3)&= 2 \binom{\frac{\lambda}{2}}{1} \frac{\alpha}{\mu} \cdot \Bigg(1-\frac{\alpha+\beta_1}{\mu}\Bigg)^2 \varphi_3(k) \\
& \approx \frac{\lambda \alpha}{\mu}\Bigg(1-\frac{\alpha}{\mu}\Bigg)^2 \varphi_3(k) \\
P(E_4)&=2 \binom{\frac{\lambda}{2}}{1}\frac{\beta_1}{\mu}\Bigg(1-\frac{\alpha}{\mu}\Bigg) \Bigg(1-\frac{\alpha + \beta_1}{\mu}\Bigg)^2 \varphi_4(k)\\
& \approx \frac{\lambda}{\mu}\Bigg(1-\frac{\alpha}{\mu}\Bigg)^3 \varphi_4(k)
\end{align*} 
The probability to evolve one more elite species is ($P(E_1), P(E_2)$ are the same as in the previous derivation):  
\begin{align*}
P(S_{\alpha,k})=P(E_1)+P(E_2)+P(E_3)+P(E_4)
\end{align*}
and the expected time until there are $\delta \mu$ elite strings in the population:
\begin{align}
\label{eq:fht2}
\mathbf{E}\tilde{T}_{\mu,k} & \leq \sum_{\alpha=1}^{\delta \mu}\frac{1}{S_{\alpha,k}} \nonumber \\
&=2 \mu^4 \sum_{\alpha=1}^{\delta \mu} \frac{1}{b_3 \alpha^3 +b_2 \alpha^2 +b_1 \alpha +b_0} \nonumber \\
&=\frac{2 \mu^4}{b_3} \sum_{\alpha=1}^{\delta \mu} \frac{1}{\alpha^3 +\frac{b_2}{b_3} \alpha^2 +\frac{b_1}{b_3} \alpha +\frac{b_0}{b_3}}
\end{align} 
where 
\begin{align}
b_0&=\lambda \mu^2(2 \mu \varphi_4(k)+\varphi_2(k)) \nonumber \\
b_1&=2 \lambda \mu(\mu \varphi_1(k)+\mu^2 \varphi_3(k)-3 \mu \varphi_4(k) -\varphi_2(k)) \nonumber \\
b_2&=\lambda(\varphi_2(k)-4 \mu^2 \varphi_3(k)-2 \mu \varphi_1(k)-6 \mu \varphi_4(k)) \nonumber \\
b_3&=2 \lambda(\mu \varphi_3(k)-\varphi_4(k)) \nonumber
\end{align}
Full solution of Equation \ref{eq:fht2} is in Appendix \ref{sec:eq6}.\\
\linebreak
The upper bound on expected optimization time is (for $\delta \neq \frac{c}{\mu}, c$ is a constant):
\begin{equation}
\mathbf{E}\tau_{(\mu + \lambda)EA_{1BS}} = \frac{c\mu^{1+\epsilon} n \log n}{\lambda} - O \bigg( \frac{\mu^{1+\epsilon} n}{\lambda} \bigg)
\end{equation}
for $0 < \epsilon(\mu) <1$ and if $\delta = \frac{c}{\mu}$, in the number of function evaluations:  
\begin{equation}
\mathbf{E}\tau_{(\mu + \lambda)EA_{1BS}} = c \mu n \log n - O(\mu n)
\end{equation}
This bound is sharper than the one obtained using simpler arguments earlier in this paper up to the order $\lambda$ (since more possibilities of adding elite species are considered). It is also comparable to the results in \cite{chenhe09, jansen05, he2010} (see below). Such a result likely means that population has positive effect for some relatively small $\mu$, but as it keeps increasing it either levels out (at best) or starts to degrade performance. 	
\subsection{Generations vs Function evaluations}
\noindent Tournament selection has a property that you do not need to evaluate every species, but we need to make $2 \lambda$ evaluations (since two species compete for 1 slot in the recombination pool, so the number of evaluations each generation is $O(\lambda)$. Therefore, in terms of the number of functions evaluations the rough bound becomes $O(\mu \lambda n \log n)$ and the refined one $O(\mu n \log n)$. If $\mu=\lambda=O(1)$ this reduces to the well-known result of $O(n \log n)$ for OneMax function. The $\lambda$ term in the denominator means that for the algorithm run on parallel computers the increase in the recombination pool size improves the performance. 
\subsection{Comparison to earlier results}
\noindent The closest comparison we can draw is to ($N+N$)EA with mutation and tournament selection function in \cite{chenhe09}, $O(n N \log N + n \log n)$ if measured in the number of function evaluations (Proposition 4). By setting $N=O(1)=c \geq 1$ this bound becomes $n \log n+O(n)$, which is larger than just $O(n \log n)$. If instead we set $\mu=N=O(\sqrt{\log{n}}) \textnormal{ or } O(\frac{\log n}{\log \log n})$ the result in \cite{chenhe09} is sharper than in this paper. For populations $\Omega(\frac{\sqrt{n}}{\log n})$ though the bound in this article becomes sharper again, e.g., for $\mu=N=O(\sqrt{n})$ it is $cn^{\frac{3}{2}}\log n-n^{\frac{3}{2}}$, and in \cite{chenhe09} it is $\frac{n^{\frac{3}{2}} \log n}{2}+O(n \log n)$.
\section{Discussion}
\noindent We presented a new tool to analyze population-based elitist EAs, Elitism Levels Traverse Mechanism, which we used to derive a new upper bound on $(\mu+\lambda)$ EAs with a recombination operator and a variant of tournament selection solving OneMax problem.\\
\linebreak
We derived and proved the lower bound on the probability of evolving exactly 1 new currently-elite species, which helped us obtain the upper bound on the expected optimization time.\\ 
\linebreak
We showed that for a function with fitness plateaus it is harder to add a super-elite offspring to the population than an elite offspring for a function without plateaus. This means that the very number of super-elite species in the population is more important in the former case than the number of elite species in the latter.\\
\linebreak
It may seem from the derived equations that population generally degrades performance (since $\mu$ is in the numerator), but for small size of population, when the cost of functions evaluations is not much different from 1, population brings about some positive effect.\\
\linebreak
As it keeps increasing, the effect levels out, at the same time the costs of evaluating functions grows and population loses its benefit. For other algorithms, s.a. RLS the effect even of small-sized population is usually negative, which makes EA+1BS (and, possibly, other recombination-based algorithms) stand out.\\
\linebreak
At the same time the recombination pool improves performance (at least when measured in terms of the number of generations), since $\lambda$ is in the denominator. This means there is a benefit from increasing recombination pool size when the algorithm is run on parallel computers.\\
\linebreak
The Mechanism we have designed in this article proved to be quite efficient in deriving upper bounds for OneMax function and we are confident it can also yield tight upper bounds on other population-driven algorithms and more complicated problems.
\section{Conclusions and Future Work}
\noindent There are many reasons to use population in evolutionary computing rather than just $(1+1) \textnormal{ or } (1,1)$ algorithms, that includes higher diversity and shorter evolutionary path (see \cite{chenhe09}). We intend to expand the results in this article by considering the following extensions to the upper bound tool:
\begin{enumerate}
\item Analysis of functions with fitness plateaus. Apparently for functions with fitness plateaus (e.g. Royal roads) both large populations and large number of elite parents are crucial compared to functions without one, so we will extend our findings to these functions as well. 
\item Typical runtime analysis. It is fairly obvious that the actual number of elite species grows every generation at some rate that realistically lies between the upper and lower bounds. We need to find an approximation on the expected number of $\alpha$ added to the population every generation and thus estimate the typical runtime.
\item Elitism rates analysis. In this article we never really considered the rate of elitism, i.e. the actual number of species saved in the population each generation, although numerical computation shows that it has a strong effect on the runtime. So far we only said that all the elite species are saved each generation, thus accumulating over time till $\delta \mu$. It would be interesting to compare elitism level 1 to 50\%, i.e. if there is any difference if only 1 species is saved compared to half of the population.      
\item Derivation of $\delta \mu$ to find the proportion of elite species that yields a high enough probability of evolution. Quite obviously it is different for functions with plateaus and without.
\item Derivation of the optimal population size. We will do this by comparing the number of functions evaluations necessary of $(1+1) \textnormal{ and } (\mu + \lambda)$ algorithms. 
\end{enumerate}
\bibliography{IEEEabrv,mybib9}
\bibliographystyle{IEEEtran}
\appendices
\section{Proof of  Equations \ref{eq:main_ineq}-\ref{eq:main_eq}}
\label{sec:main_eq}
\noindent Main idea and logic of the lower bound on the probability of adding an elite offspring and the upper bound on runtime following from this is presented in Section \ref{sec:main_sec}. Here we present the derivation of this bound.\\
\linebreak
We prove this lower bound inequality for an arbitrary subset (it is not to be confused with at trivial one of the form $\sum_{k \geq r} \binom{n}{k}p^k (1-p)^k > \binom{n}{r}p^r (1-p)^{n-r}$):
\begin{align*}
P(E_j) &= \binom{\frac{\lambda}{2}}{1}P_{sel}P_{swap} = \binom{\frac{\lambda}{2}}{1}P_{sel} \varphi_j(k) \\ 
P(E^{\ast}_j) &= P \Bigg( \bigcup_{r=1}^{m_j}E^{\ast}_{jr} \Bigg) \\
&=\sum_{r=1}^{m_j}\binom{\frac{\lambda}{2}}{r}P_{sel}^r (1-P_{sel})^{\frac{\lambda}{2}-r} \binom{r}{1} \varphi_j(k)(1-\varphi_j(k))^{r-1} \\
\end{align*} 
In this expression $P_{swap}$ is not necessarily the probability to swap bits $<0,1>$. It is the probability to swap bits such that an elite offspring evolves. Since all the terms in the sum are positive, we use the lower bound on this expression:
\begin{align*}
P(E^{\ast}_j) & \geq \sum_{r=1}^{m_j}\binom{\frac{\lambda}{2}}{r}P_{sel}^r (1-P_{sel})^{\frac{\lambda}{2}-r} \varphi_j(k)(1-\varphi_j(k))^{r-1} \\
& \geq  \frac{\varphi_j(k)}{(1-\varphi_j(k)} \Bigg(\sum_{r=0}^{m_j}\binom{\frac{\lambda}{2}}{r}P_{sel}^r(1-\varphi_j(k))^r (1-P_{sel})^{\frac{\lambda}{2}-r} \\
&- (1-P_{sel})^{\frac{\lambda}{2}}  \Bigg) \\
& \geq \varphi_j(k) \Bigg(\sum_{r=0}^{m_j}\binom{\frac{\lambda}{2}}{r}P_{sel}^r(1-\varphi_j(k))^r (1-P_{sel})^{\frac{\lambda}{2}-r} - \\
&(1-P_{sel})^{\frac{\lambda}{2}}  \Bigg) 
\end{align*}
Canceling out $\varphi_j(k)$ and moving the term $ e^{-1} \leq (1-P_{sel})^{\frac{\lambda}{2}} \leq \frac{1}{\sqrt{e}} < 1$ on the other side, LHS of the inequality becomes 
\begin{align*}
P(E^{\ast}_j) &\geq \sum_{r=0}^{m_j}\binom{\frac{\lambda}{2}}{r}P_{sel}^r(1-\varphi_j(k))^r (1-P_{sel})^{\frac{\lambda}{2}-r} \\
& \geq (P_{sel}(1-\varphi_j(k))+1-P_{sel})^{\frac{\lambda}{2}} \\
& = (1-P_{sel} \varphi_j(k))^{\frac{\lambda}{2}}
\end{align*}
and the RHS is upper-bounded by 
\begin{equation*}
\frac{1}{\sqrt{e}} + \frac{\lambda P_{sel}}{2} = \frac{1}{\sqrt{e}} + o(\lambda^{c-1}) \textnormal {by the argument below}
\end{equation*}
LHS is lower-bounded by (using Bernoulli inequality for $\frac{\lambda}{2} \geq 1$):
\begin{align*}
P(E^{\ast}_j) &\geq (1-P_{sel}\varphi_j(k))^{\frac{\lambda}{2}} \geq 1-\frac{\lambda P_{sel} \varphi_j(k)}{2} 
\end{align*}
Since we can select $P_{sel}=o(\lambda^c)$ and $\varphi_j(k)=O(\frac{1}{n^{c}}) , c \in \mathbb{Z}$, the expression is
\begin{equation}
P(E^{\ast}_j)=1-o(1) > \frac{1}{\sqrt{e}}+o(1)=P(E_j)
\end{equation}
thus proving the upper bound on the probability of evolving 1 more elite species for an arbitrary subset. This logic applies for each of the M subsets (types of pairs) of the recombination pool, and the inequality becomes
\begin{equation}
P \Bigg( \bigcup_{j=1}^{M} E^{\ast}_j \Bigg) > \sum_{j=1}^{M} P(E_j)
\end{equation}   
The upper bound in Equation \ref{eq:main_eq} follows directly. 
\linebreak
\section{Solution of Equation \ref{fht1}}
\label{sec:eq2}
We have a quadratic equation
\begin{equation*}
S(\mu,k) = \sum_{\alpha} P_2(\alpha)=\sum_{\alpha=1}^{\delta \mu}\frac{1}{\alpha^2 +b_1 \alpha +b_0}
\end{equation*}
with 
\begin{align*}
b_0 &= \frac{\mu^2 \varphi_2(k)}{\varphi_2(k)-2 \mu \varphi_1(k)} \\
b_1 &= \frac{2 \mu(\mu \varphi_1(k)-\varphi_2(k))}{\varphi_2(k) -2 \mu \varphi_1(k)}
\end{align*}
In order to simplify the already complicated derivation, we want the expression above in the form 
\begin{equation*}
P_2(\alpha)=\frac{1}{(\alpha+r)^2} =\frac{1}{(\alpha^2 +2 r \alpha +r^2)}
\end{equation*}
for some $r$, not necessarily rational. From equating coefficients it becomes clear that 
\begin{equation*}
r=\sqrt{b_0} \lor r=\frac{b_1}{2}
\end{equation*} 
and so, using the first root 
\begin{align*}
S(\mu,k)=\sum_{\alpha=1}^{\delta \mu}\frac{1}{(\alpha+r)^2} = \psi_{1}(\sqrt{b_0}) - \psi_{1}(\sqrt{b_0} + \delta \mu + 1)
\end{align*}
For large $b_0$ these expressions involving digamma function can be expanded asymptotically in Taylor series (we use only the first two terms):
\begin{align*}
S(\mu,k) &\approx \Bigg( \frac{1}{b_0} - \frac{1}{2b_0} \Bigg)-\Big( \frac{1}{b_0} -\frac{\delta \mu}{b_0} - \frac{1}{2 b_0} \Big) = \frac{\delta \mu}{b_0} \\
&=\frac{\delta \mu(\varphi_2(k)-2 \mu \varphi_1(k))}{\mu^2 \varphi_2(k)} = \frac{\delta (\varphi_2(k)-2 \mu \varphi_1(k))}{\mu \varphi_2(k)}
\end{align*}
and therefore the expected time to traverse enough levels of elitism to improve 1 bit of the string is (plugging this expression into Equation \ref{fht1}) 
\begin{align*}
\mathbf{E}\tilde{T}_{\mu,k} &= \frac{2 \mu^4}{\lambda (\varphi_2(k)-2 \mu \varphi_1(k))} \cdot  \frac{\delta (\varphi_2(k)-2 \mu \varphi_1(k))}{\mu \varphi_2(k)} \\
&=\frac{2 \mu^3 \delta}{\lambda \varphi_2(k)}
\end{align*}
To improve the pair $<\beta_1, \beta_1>$ we need to either swap 1 from the first parent and 0 from the second, or the other way around (any other outcome just keeps the current number of bits in each parent):
\begin{equation*}
\varphi_2(k) = 2 \cdot \frac{k-1}{n} \cdot {\frac{n-k+1}{n}} =\frac{2(k-1)(n-k+1)}{n^2}
\end{equation*}
Plugging this into the expression for $\mathbf{E}\tilde{T}_{k}$, we obtain the expected optimization time of the algorithm, pessimistically assuming that at the beginning of the run the best species has only 2 1-bit and finishes at $n-2$, since if the fitness of $\beta_1 = n-1$ implies the fitness of $\alpha = n$. 
\begin{align*}
\mathbf{E}\tau_{(\mu+\lambda)EA_{1BS}} &\leq \frac{\mu^3 n^2 \delta}{\lambda}\sum_{k=2}^{n-2}\frac{1}{(k-1)(n-k+1)} \\
&=\frac{\delta \mu^3 n^2}{\lambda}\cdot \frac{1}{n}\Bigg(\sum_{k=2}^{n-2}\frac{1}{k-1} + \sum_{k=2}^{n-2}\frac{1}{n-k+1}\Bigg)\\
&=\frac{\delta \mu^3 n}{\lambda} \Bigg(\log (n-1) + O(1) \Bigg)
\end{align*}
The second step is due to partial fraction expansion. Although this seems quite a loose bound given cubic in $\mu$, we take $\mu=O(\lambda)$ so all we need to establish is $\delta$ to reduce the power.\\
\linebreak
Obviously $0 <\delta <1$, but we need to select it s.t. summation over $\alpha$ makes sense. We set $\delta = \mu^{-\varepsilon_1}$ for an arbitrary $\varepsilon_1 >0 \textnormal { s.t. } \delta \mu=\mu^{1-\varepsilon_1} >1$. Then $\delta \mu^2 = \mu^{2-\varepsilon_1}=\mu^{1+\varepsilon_2}$. For example, $\varepsilon=\frac{1}{2} \textnormal { yields } \delta \mu = \sqrt{\mu} \textnormal { and } \delta \mu^2 = \sqrt{\mu^3}$. Therefore, the upper bound on the expected convergence time is 
\begin{equation}
\mathbf{E}\tau_{(\mu+\lambda)EA_{1BS}} = O(\mu^{1+\varepsilon_2} n \log n)
\end{equation}     
In fact if (similar to \cite{chenhe09}) we set $\delta=\frac{c}{\mu} \textnormal{ for } c \in \mathbb{Z^{+}}, \textnormal{ we get } \delta \mu=c \textnormal{ and } \delta \mu^2 =c \mu = O(\mu)$, so the expectation becomes linear in $\mu$:
\begin{equation}
\mathbf{E}\tau_{(\mu+\lambda)EA_{1BS}} = O(\mu n \log n)
\end{equation}   
\section{Solution to Equation \ref{eq:fht2}}
\label{sec:eq6}
We need a solution to the cubic equation of the form
\begin{align*}
S(\mu, k)&=\sum_{\alpha}P_3(\alpha)= \sum_{\alpha=1}^{\delta \mu} \frac{1}{\alpha^3 + b'_2 \alpha_2 +b'_1 \alpha +b_0}
\end{align*}
where 
\begin{align*}
b'_2 &= \frac{\varphi_2(k)-4 \mu^2 \varphi_3(k) -2 \mu \varphi_1(k) - 6 \mu \varphi_4(k)}{2(\mu \varphi_3(k)-\varphi_4(k))}\\
b'_1 &= \frac{\mu(\mu \varphi_1(k)+\mu^2 \varphi_3(k)-3 \mu \varphi_4(k) -\varphi_2(k))}{2 (\mu \varphi_3(k)-\varphi_4(k))}\\
b'_0 &= \frac{\mu^2(2 \mu \varphi_4(k)+\varphi_2(k))}{2 (\mu \varphi_3(k)-\varphi_4(k))} \\
\end{align*}
Solution to $S(\mu, k)$ is of the form 
\begin{equation*}
S(\mu,k)=\sum_{\alpha}\frac{1}{(\alpha+\rho)^3}=\sum_{\alpha}\frac{1}{\alpha^3 + 3 \alpha^2 \rho + 3 \alpha \rho^2 +\rho^3}
\end{equation*}
Equating the coefficients we obtain three roots $\rho$:
\begin{align*}
\rho &= \frac{b'_2}{3} \\
\rho &= \pm \frac{\sqrt{b'_1}}{3}\\
\rho &= \sqrt[3]{b'_0}
\end{align*} 
To simplify the increasingly hard notation, we select only the last root:
\begin{align*}
S(\mu,k) &=\sum_{\alpha=1}^{\delta \mu} \frac{1}{(\alpha+\sqrt[3]{b'_{0}})^3} =\frac{\psi_{2}(\sqrt[3]{b'_{0}}+ \delta \mu +1) - \psi_{2}(\sqrt[3]{b'_{0}}+ 1)}{2}\\
&= \frac{1}{2} \Bigg(\Bigg(-\frac{1}{\sqrt[3]{{b'_{0}}^{2}}}+\frac{2 \delta \mu +1}{b'_0} \Bigg) - \Bigg(-\frac{1}{\sqrt[3]{{b'_{0}}^{2}}} + \frac{1}{b'_0} \Bigg) \Bigg) \\
&=\frac{1}{2} \cdot \frac{2 \delta \mu}{b'_0} = \frac{\delta \mu}{b'_0}
\end{align*}
The second line in the derivation was obtained by expanding both second-order polygamma functions in Taylor series as $b'_{0} \to \infty$ and taking two first terms of each function. We now combine the front term in Equation \ref{eq:fht2} with this derivation to obtain the expression on the upper bound on achieving the number of elite species in the population $\delta \mu$:
\begin{equation*}
\mathbf{E}\tilde{T}_{\mu,k}  \leq \frac{2 \mu^5 \delta}{\lambda b_3 b'_0} = \frac{2 \delta \mu^3}{\lambda(2 \mu \varphi_3(k)+\varphi_4(k))}
\end{equation*}   
since 
\begin{align*}
b_3 b'_0 &= 2 (\mu \varphi_3(k)-\varphi_4(k)) \cdot \frac{\mu^2(2 \mu \varphi_3(k)+\varphi_4(k))}{2 (\mu \varphi_3(k)-\varphi_4(k))} \\
&=\mu^2(2 \mu \varphi_3(k)+\varphi_4(k))
\end{align*}
We are now ready to find the upper bound on the expected optimization time of the algorithm:
\begin{equation}
\label{eq:fht_aux}
\mathbf{E}\tau_{(\mu + \lambda)EA_{1BS}} \leq \sum_{k=3}^{n-3}\mathbf{E}\tilde{T}_{\mu, k}= \frac{2 \delta \mu^3}{\lambda} \sum_{k=3}^{n-1}\frac{1}{2 \mu \varphi_3(k)+\varphi_4(k)}
\end{equation}
Here again we pessimistically assume that the best species at the start of the run has fitness 3, since in such case fitness of $\beta_{-1}$ has minimal fitness of 1, otherwise we obtain inconsistencies s.a. $\frac{1}{0}$. We have two probabilities to consider for the two new types of pairs:
\begin{align*}
<\alpha ,\beta_{-1}>: \varphi_3(k) &= \frac{k}{n} \cdot \frac{k-2}{n} + \frac{n-k}{n} \cdot \frac{n-k+2}{n} \\
&=\frac{k(k-2)+(n-k)(n-k+2)}{n^2}
\end{align*}
We need to preserve the better parent in order to get it added to the population, so need to either select 1-bits in each parent or 0-bits in each parent. for the last swap probability, $\varphi_4(k)$, we need only to select a 0 in the $\beta_1$ parent and 1 in $\beta_{-1}$ parent, other options either degrade the better parent or leave the current fitness.
\begin{equation*}
<\beta_1, \beta_{-1}>: \varphi_4(k)= \frac{(n-k+1)(k-2)}{n^2}
\end{equation*}
We continue with manipulating with the summand over $k$:
\begin{align*}
&S(n,k) = \frac{1}{2 \mu \varphi_3(k)+\varphi_4(k)}\\ 
&= \frac{n^2}{(4 \mu-1)k^2+(n-8 \mu -4 \mu n +3)k +4 \mu n -2n+2 \mu n^2-2}\\
& \leq \frac{n^2}{\mu} \cdot \frac{1}{k^2-4(n+2)k+2n(n+1)}
\end{align*}
We leave out the first fraction, and factor the denominator in the form $(k-s)(k-r)$, s.t. $s,r$ are solutions to the set of equations:
\begin{equation*}
\Bigg\{\left.
\begin{array}{ll}
s+r &= 4(n+2)\\
sr&=2n(n+1)
\end{array}
\right.
\end{equation*} 
The resulting solution (we only use one of the roots, which are symmetric) is:
\begin{equation*}
\Bigg\{\left.
\begin{array}{ll}
s &= 2n +\sqrt{2} \sqrt{n^2+7n+8}+4 \\
r &= 2n - \sqrt{2} \sqrt{n^2+7n+8}+4
\end{array}
\right.
\end{equation*}
The value under the root can be bounded by
\begin{equation*}
n+2 \leq  \sqrt{n^2+7n+8} \leq n+4 
\end{equation*}
So the expression becomes upper-bounded by 
\begin{equation*}
\frac{1}{(k-(2n+\sqrt{2}(n+2)))(k-(2n-\sqrt{2}(n+4)))}
\end{equation*}
Expanding this in partial fractions, we obtain 
\begin{align*}
\frac{1}{2\sqrt{2}(n+3)} &\cdot \Bigg(\frac{1}{k-(2n+\sqrt{2}(n+2))} \\
&- \frac{1}{k-(2n-\sqrt{2}(n+2))} \Bigg)
\end{align*}
We obtain two sums over k:
\begin{align*}
S_{1}(n,k)&=\sum_{k=3}^{n-3}\frac{1}{k-(2n+\sqrt{2}(n+2))} \\
&\approx \psi_{0}(n-2n-\sqrt{2}n)-\psi_{0}(3-2n-\sqrt{2}n) \\
&=\psi_{0}(-(1-\sqrt{2})n)- \psi_{0}(3-(2+ \sqrt{2})n)\\
&=O(1)-O(1)=-O(1)
\end{align*}
The result of $-O(1)$ is due to the fact that we can select any $n$, for which the values of digamma function are small negative constants (see Appendix \ref{sec:math} for details on Taylor series expansion of $\psi_{0}(n)$ for $n<0$). For the second sum, we notice the upper bound on the value in the denominator, since $2-\sqrt{2} \approx 0.58 <1$: 
\begin{align*}
&S_{2}(n,k)=\sum_{k=3}^{n-3}\frac{1}{k-(2n-\sqrt{2}(n+2))} \\
&\leq \sum_{k=3}^{n-3}\frac{1}{k-n} = - \sum_{k=3}^{n-3}\frac{1}{n-k} \approx -\log(n-3) + O(1)
\end{align*}   
the minus sign in front of the expression cancels out and we obtain the upper bound for $S(\mu,n)$:
\begin{equation*}
S(\mu,n) \leq \frac{n^2(\log(n-3)-O(1)}{\mu n}
\end{equation*} 
and the upper bound on the expected first hitting time:
\begin{equation}
\mathbf{E}\tau_{(\mu+\lambda)EA_{1BS}} \leq \frac{2 \delta \mu^2 n(\log(n-3)-O(1))}{\lambda} 
\end{equation}
with $\delta=\frac{c}{\mu}$ the expression becomes (measured in the number of generations, for $c>0$)
\begin{equation}
\mathbf{E}\tau_{(\mu+\lambda)EA_{1BS}} = \frac{c \mu n \log n}{\lambda}-O\Bigg( \frac{\mu n}{\lambda}\Bigg)
\end{equation}
or, in the number of function evaluations,
\begin{equation}
\mathbf{E}\tau_{(\mu+\lambda)EA_{1BS}}  = c \mu n \log n - O(\mu n)
\end{equation}
\section{Mathematical Expressions}
\label{sec:math}
\noindent There is a number of important mathematical expression used throughout the article, we present some of them here:
\begin{equation*}
H(n) = \sum_{k=0}^{n-1}\frac{1}{n-k} = \sum_{k=0}^{n-1}\frac{1}{k} \approx \int_{0}^{n-1}\frac{dx}{n-x} = \log n < \log n +1
\end{equation*}
Digamma function:
\begin{equation*}
\psi_{0}(n) = \log n + O \Bigg(\frac{1}{n} \Bigg)
\end{equation*}
For $\psi_0(n) \textnormal { with } n \to -\infty$ we use the largest term of Taylor series for asymptotic expansion:
\begin{equation*}
\psi_0(n) \approx \pi \cot(\pi n) + O\Bigg(\frac{1}{n}\Bigg)
\end{equation*}
The values for $\cot(\pi n)$ for integer $n$, such as in this article, are infinity. Therefore for expressions for $S_{1}(n,k) \textnormal { and } S_{2}(n,k)$ we have selected some constants, e.g. $-(1-\sqrt{2}), 2-\sqrt{2}$, s.t. the resulting values are constants. Since $n$ is arbitrarily large, we can find such $n$ that the difference between them is negative, hence we obtain term $-O(1)$.  
\end{document}